\documentclass[10pt,twocolumn,letterpaper]{article}

\usepackage{cvpr}
\usepackage{times}
\usepackage{epsfig}
\usepackage{graphicx}
\usepackage{amsmath}
\usepackage{amssymb}
\usepackage{mathtools}
\usepackage{tablefootnote}

\usepackage{IEEEtrantools}
\DeclareMathOperator*{\argmin}{arg\,min}
\usepackage{algorithm}%
\usepackage{algpseudocode}%
\usepackage{multirow}
\usepackage{CJK}

\usepackage{subfig}

\usepackage{color}

\usepackage[breaklinks=true,bookmarks=false]{hyperref}
\graphicspath{./figure}

\cvprfinalcopy 

\def\cvprPaperID{189} 

\begin{document}


\title{Scene Flow to Action Map: A New Representation for RGB-D based\\ Action Recognition with Convolutional Neural Networks}

\author{Pichao Wang$^{\rm 1}$, Wanqing Li$^{\rm 1}$, Zhimin Gao$^{\rm 1}$\thanks{Corresponding author}\thanks{To be appear in CVPR2017}, Yuyao Zhang$^{\rm 1}$, Chang Tang$^{\rm 2}$ and Philip Ogunbona$^{\rm 1}$\\
$^{\rm 1}$Advanced Multimedia Research Lab, University of Wollongong, Australia\\
$^{\rm 2}$China University of Geosciences, Wuhan, China\\
{\tt\small pw212@uowmail.edu.au,wanqing@uow.edu.au,\{zg126, yz606\}@uowmail.edu.au}\\ {\tt\small happytangchang@gmail.com, philipo@uow.edu.au}\\
}

\maketitle

\begin{abstract}
Scene flow describes the motion of 3D objects in 
real world and potentially could be the basis of a good feature for 3D action 
recognition. However, its use for action recognition, especially in the context 
of convolutional neural networks (ConvNets), has not been previously studied. 
In this paper, we propose the extraction and use of scene flow for action 
recognition from RGB-D data. Previous works have considered the depth and RGB 
modalities as separate channels and extract features for later fusion. 
We take a different approach and consider the modalities as one entity, thus 
allowing feature extraction for action recognition at the beginning.   
Two key questions about the use of scene flow for action recognition are 
addressed: how to organize the scene flow vectors and how to represent the long 
term dynamics of videos based on scene flow. In order to calculate the scene 
flow correctly on the available datasets, we propose an effective 
self-calibration method to align the RGB and depth data spatially without 
knowledge of the camera parameters. Based on the scene flow vectors, we propose 
a new representation, namely, Scene Flow to Action Map 
(SFAM), that describes several long term spatio-temporal dynamics for action 
recognition. We adopt a channel transform kernel to 
transform the scene flow vectors to an optimal color space analogous to RGB.  
This transformation takes better advantage of the trained ConvNets models over 
ImageNet. Experimental results indicate that this new representation can 
surpass the performance of state-of-the-art methods on two large 
public datasets.

\end{abstract}

\begin{figure}[t]
\begin{center}
{\includegraphics[height = 40mm, width = 85mm]{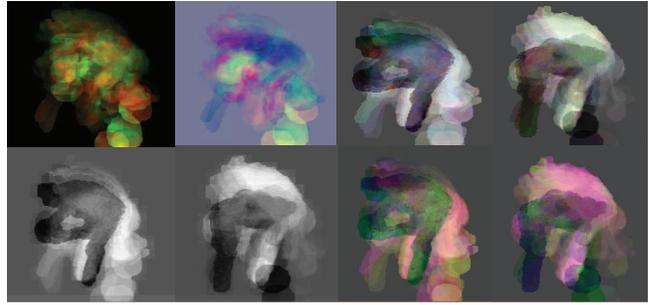}}
\end{center}
\caption{Samples of variants of SFAM for action ``Bounce Basketball" from M$^{\textbf{2}}$I Dataset~\cite{liu2016benchmarking}. For top-left to bottom-right, the images correspond to SFAM-D, SFAM-S, SFAM-RPf, SFAM-RPb, SFAM-AMRPf, SFAM-AMRPb, SFAM-LABRPf, SFAM-LABRPb. }
\label{SFAMs}
\end{figure}

\section{Introduction}
Recognition of human actions from RGB-D data has generated
renewed interest in the computer vision community due to the 
recent availability of easy-to-use and low-cost depth sensors (e.g. 
Microsoft Kinect~\texttrademark sensor). In addition to tristimulus visual data 
captured by conventional RGB cameras, depth data are provided in RGB-D cameras, 
thus encoding rich 3D structural information of the entire scene. 
Previous works~\cite{Nie2015,Kong2015CVPR,
hu2015jointly,wu2015watch,yu2016structure,jia2016low} showed the effectiveness 
of fusing the two modalities for 3D action recognition. However, all the 
previous methods consider the depth and RGB modalities as separate channels 
from which to extract features and fuse them at a later stage for action 
recognition. Since the depth and RGB data are captured simultaneously, it will 
be interesting to extract features considering them jointly as one entity. 
Optical flow-based methods for 2D action 
recognition~\cite{wang2013action,lan2015beyond,peng2014action,peng2016bag, 
wang2015action} have provided the state-of-the-art results for several years. 
In contrast to optical flow which provides the projection of the scene motion 
onto the image plane,
scene flow~\cite{vedula2005three,hadfield2014scene,menze2015object,hornacek2014sphereflow,jaimezprimal,sun2015layered,quiroga2014dense}
estimates the actual 3D motion field. 
Thus, we propose the use of scene flow for 3D action recognition.  
Differently from the optical flow-based late fusion methods on RGB and depth 
data, scene flow extracts the real 3D motion and also explicitly preserves 
the spatial structural information contained in RGB and depth modalities.

There are two critical issues that need to be addressed when adopting 
scene flow for action recognition: how to organize the scene flow vectors and 
how to effectively exploit the spatio-temporal dynamics. 
Two kinds of motion representations can be identified: 
Lagrangian 
motion~\cite{wang2013action,lan2015beyond,peng2014action,
wang2015action,peng2016bag,pichao2016} and Eulerian 
motion~\cite{bobick2001recognition,
man2006individual,Yang2012a,pichao2015,pichaoTHMS,bilen2016dynamic}. Lagrangian 
motion focuses on individual points and analyses their change in location over 
time. Such trajectories requires reliable point tracking over long 
term and is prone to error. Eulerian motion considers a set of 
locations in the image and analyses the changes at these locations over time, 
thus avoiding the need for point tracking.

Since scene flow vectors could be noisy and to avoid the difficulty of long 
term point tracking of Lagrangian motion, we adopted the Eulerian approach in 
constructing the final representation for action recognition. Furthermore, the 
scene flow between two consecutive pair of RGB-D frames (two RGB images and two 
corresponding depth images) is one simple Lagrangian motion with only two 
frames matching/tracking. This property provides a better representation than 
is possible with Eulerian motion obtained from raw pixels.

However, it remains unclear as to how video could be effectively represented 
and fed to deep neural networks for classification. For example, one can 
conventionally consider a video as a sequence of still images with some form of 
temporal smoothness, or as a subspace of images or image features, or as the 
output of a neural network encoder. Which one among these and other 
possibilities would result in the best representation in the context of action 
recognition is not well understood. The promising performance of existing 
temporal encoding 
works~\cite{pichao2015,pichaoTHMS,pichao2016,bilen2016dynamic} provides a 
source of motivation. These works encode the spatio-temporal information as 
dynamic images and enable the use of existing ConvNets models directly without 
training the whole networks afresh. Thus, we propose to encode the RGB-D video 
sequences based on scene flow into one motion map, called Scene Flow to Action 
Map (SFAM), for 3D action recognition. Intuitively and similarly to the three 
channels of color images, the three elements of a scene flow vector can be 
considered as three channels. Such consideration allows the scene flow between 
two consecutive pairs of RGB-D frames to be reorganized as one three-channel 
Scene Flow Map (SFM), and the RGB-D video sequence can be represented as SFM 
sequence.  In the spirits of Eulerian motion and rank pooling 
methods~\cite{fernando2016rank,bilen2016dynamic}, we propose to encode SFM 
sequence into SFAM.  Several variants of SFAM are developed. They capture the 
spatio-temporal information from different perspectives and are complementary 
to each other for final recognition.  However, two issues arise with these 
hand-crafted SFAMs: 1) direct organization of the scene flow vectors in SFM may 
sacrifice the relations among the three elements; 2) in order to take 
advantage of available model trained over ImageNet, the input needs to be 
analogous to RGB images; that is, the input for the ConvNets need to have 
similar properties to conventional RGB images as used in trained 
filters. Based on these two observations, we propose to learn Channel Transform 
Kernels  with  rank pooling method and ConvNets, that convert the 
three channels into suitable three new  channels capable of exploiting the 
relations among the three elements and have similar RGB image features.  
With this transformation, the dynamic SFAM can describe both 
the spatial and temporal information of a given video. It can be used as the 
input to available and already trained ConvNets along with fine-tuning.

The contributions of this paper are summarized as follows:1) 
The proposed SFAM is the first attempt, to our best knowledge, to extract 
features from depth and RGB modalities as joint entity through scene flow, 
in the context of ConvNets; 2) we propose an effective self-calibration method 
that enables the estimation of scene flow from unregistered captured RGB-D data; 
3) several variants of SFAM that encode 
the spatio-temporal information from different aspects and are complementary 
to each other for final 3D action recognition are proposed;
4) we introduce Channel Transform Kernels which learn the relations among the 
three channels of SFM and convert the scene flow vectors to RGB-like images to 
take advantages of trained ConvNets models and 5) the proposed method achieved 
state-of-the-art results on two relatively large datasets.

The reminder of this paper is organized as follows. Section~\ref{relatedwork} 
describes the 
related work. Section~\ref{SFAM} introduces the SFAM and its variants, and 
presents the proposed Channel Transform Kernels. Experimental results on two 
datasets are provided in Section~\ref{results}. Section~\ref{conclusion} 
concludes the paper and discusses future work.

\begin{figure*}[t]
\begin{center}
{\includegraphics[height = 32mm, width = 160mm]{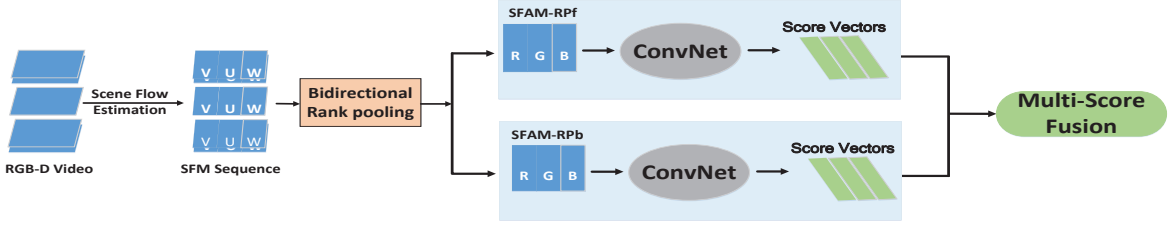}}
\end{center}
\caption{Illustration of Multiply-Score Fusion for SFAM-RP. }
\label{scorefusion}
\end{figure*}

\section{Related Work}\label{relatedwork}

\subsection{Feature Extraction from RGB-D Data}
 Since the first work~\cite{li2010action} on 3D action recognition from depth 
data captured by commodity depth sensors (e.g., Microsoft Kinect~\texttrademark) 
in 2010, many 
methods for action recognition have been proposed based on depth, RGB or skeleton data. 
These methods either extracted features from one modality: 
depth~\cite{wang2012mining,Yang2012a,Oreifej2013,yangsuper,lurange,pichao2015,
pichaoTHMS} or RGB~\cite{ni2015pose,rahmani2015learning} or 
skeleton~\cite{vemulapalli2014human,pichao2014,pichao2016,du2015hierarchical,shahroudy2016ntu,pichaoSPL}, 
or fuse the features extracted separately from them at a later 
stage~\cite{Nie2015,Kong2015CVPR,hu2015jointly,
wu2015watch}. Neither of these methods considered 
depth and RGB modalities jointly in feature extraction. In contrast, we propose 
to adopt scene flow for 3D action recognition and extract features jointly from 
RGB-D data.

\subsection{Scene Flow}
In general, scene flow is defined as the dense or semi-dense non-rigid motion
field of a scene observed at different instants of 
time~\cite{vedula2005three,sun2015layered,quiroga2014dense}. The 
term ``scene flow" was firstly coined by Vedula et al.~\cite{vedula2005three} 
who proposed to start by computing the Lucas-Kanade optical flow and applied 
the range flow constraint equation at a later stage. Since this work, several 
methods~\cite{zhang20013d,
wedel2011stereoscopic,vogel2013piecewise} have been proposed based 
on stereo or multiple view camera systems.
With the advent of affordable RGB-D cameras, scene flow methods for
RGB-D data have also been 
proposed~\cite{
vedula2005three,hornacek2014sphereflow,
sun2015layered,quiroga2014dense}. However most of the existing methods 
incur high computational burden, taking from several seconds to a few hours to 
compute the scene flow per frame. Thus, limiting their usefulness in real 
applications. Recently, a primal-dual framework for real-time dense RGB-D 
scene flow~\cite{jaimezprimal} has been proposed. A primal-dual algorithm is 
applied to solve the variational formulation of the scene flow problem. It is 
an iterative solver performing pixel-wise updates and can be efficiently 
implemented on GPUs. In this paper, we used this algorithm for scene flow 
calculation. 

\subsection{Deep Leaning based Action Recognition}\label{dl4ar}
Existing deep learning approaches for action recognition can be generally 
divided into four categories based on how the video is represented and fed to a 
deep neural network. The first category views a video either as a set of still 
images~\cite{yue2015beyond} or as a short and smooth transition between similar 
frames~\cite{simonyan2014two}, and each color channel of the images is fed to 
one channel of a ConvNet. Although obviously suboptimal, considering the video 
as a bag of static frames performs reasonably well. The second category 
represents a video as a volume and extend ConvNets to a third, temporal 
dimension~\cite{ji20133d,tran2015learning} replacing 2D filters with 3D 
equivalents. So far, this approach has produced little benefits, probably due to 
the lack of annotated training data. The third category treats a video as 
a sequence of images and feed the sequence to an Recurrent Neural Network 
(RNN)~\cite{donahue2015long,du2015hierarchical,veeriah2015differential,shahroudy2016ntu,liu2016spatio,junliu2017}. An RNN is typically considered as memory 
cells, which are sensitive to both short as well as long term patterns. It 
parses the video frames sequentially and encodes the frame-level information in 
the memory. However, using RNNs did not give an improvement over temporal 
pooling of convolutional features~\cite{yue2015beyond} or even over hand-crafted 
features. The last category represents a video in one or multiple compact 
images and adopt available trained ConvNet architectures for 
fine-tuning~\cite{pichao2015,pichaoTHMS,pichao2016,bilen2016dynamic,hou2016skeleton,pichaoicprwb,pichaoicprwa}. This 
category has achieved state-of-the-art results in action recognition on many 
RGB and depth/skeleton datasets. The proposed method in this paper falls into 
this last category.

\section{Scene Flow to Action Map}\label{SFAM}

SFAM encodes the dynamics of RGB-D sequences based 
on scene flow vectors. 
To make our description self-contained, in Section~\ref{pdflow} we briefly 
present the primal-dual framework for real-time dense RGB-D scene flow 
computation (hereafter denoted by PD-flow~\cite{jaimezprimal}). For scene flow 
computation, we assume that the depth and RGB data are prealigned. If this is 
not the case, the videos can be quickly realigned as described in 
Section~\ref{selfcalibration}. Then, in Section~\ref{sfam} we present several 
hand-crafted constructions of SFAM  and we propose an end-to-end learning 
method for SFAM through Channel Transform Kernels in 
Section~\ref{sfam-learning}.

\subsection{PD-flow}\label{pdflow}

The PD-flow estimates the dense 3D motion field of a scene between 
two instants of time $t$ and $t+1$ using RGB and depth images provided by an 
RGB-D camera. This motion field $\textbf{M}: (\Omega \in \mathbb{R}^{2}) 
\rightarrow \mathbb{R}^{3}$  defined over the image domain $\Omega$, is 
described with respect to the camera reference frame and expressed in meters per 
second. For simplicity, the bijective relationship $\Gamma: \mathbb{R}^{3} 
\rightarrow \mathbb{R}^3$ between $\textbf{M}$ and  $\textbf{s} = 
(\mu,\upsilon,\omega)^{T}$ is given by:
\begin{equation}
\textbf{M} = \Gamma(\textbf{s}) = \begin{pmatrix}
 \frac{Z}{f_{x}} & 0 & \frac{X}{Z} \\     
 0 & \frac{Z}{f_{y}} & \frac{Y}{Z}\\ 
 0 & 0 & 1 \\ 
\end{pmatrix}
\begin{pmatrix}
 \mu\\
\upsilon\\     
\omega\\     
\end{pmatrix},
\end{equation}
where $\mu,\upsilon$ represent the optical flow and $\omega$ denotes the range 
flow; $f_{x}, f_{y}$ are the camera focal length values, and $X,Y,Z$ the 
spatial coordinates of the observed point. Thus, estimating the optical and 
range flows is equivalent to estimating the 3D motion field but leads to a 
simplified implementation. In order to compute the motion field a minimization 
problem over $\textbf{s}$ is formulated where photometric and geometric 
consistency are imposed as well as a regularity of the solution:
\begin{equation}\label{energyfunc}
\min\limits_{\textbf{s}}\{E_{D}(\textbf{s}) + E_{R}(\textbf{s})\}.
\end{equation}
In Eq.~(\ref{energyfunc}), $E_{D}(\textbf{s})$ is the data term, representing a 
two-fold restriction for both intensity and depth matching between pairs of 
frames; $E_{R}(\textbf{s})$ is the regularization term which both smooths the 
flow field and constrains the solution space.

For data term $E_{D}(\textbf{s})$, the $L_{1}$ norm of photometric consistency 
$\rho_{I}(\textbf{s},x,y)$ 
and geometric consistency $\rho_{z}(\textbf{s},x,y)$ is minimized as:
\begin{equation}
E_{D}(\textbf{s}) = \int|\rho_{I}(\textbf{s},x,y)| + \varepsilon(x,y)|\rho_{z}(\textbf{s},x,y)|dxdy,\label{dt}
\end{equation}
where $\varepsilon(x,y)$ is a positive function that weights geometric consistency against brightness constancy; $\rho_{I}(\textbf{s},x,y) = I_{0}(x,y) - I_{1}(x+\mu,y+\upsilon)$ and $\rho_{z}(\textbf{s},x,y) = \omega - Z_{1}(x+\mu,y+\upsilon) + Z_{0}(x,y)$ with $I_{0}, I_{1}$ being the intensity images while $Z_{0}, Z_{1}$ the depth images taken at instants $t$ and $t+1$.

The regularization term $E_{R}(\textbf{s})$ is based on the total 
variation and takes into consideration the geometry of the scene which is 
formulated as:

\begin{gather}
E_{R}(\textbf{s}) = \lambda_{I}\int_{\Omega}|(r_{x}\frac{\partial \mu}{\partial x},r_{y}\frac{\partial \mu}{\partial y})|+|(r_{x}\frac{\partial \upsilon}{\partial x},r_{y}\frac{\partial \upsilon}{\partial y})|dxdy \notag \\
+ \lambda_{D}\int_{\Omega}|(r_{x}\frac{\partial \omega}{\partial x},r_{y}\frac{\partial \omega}{\partial y})|dxdy,\label{eq6}
\end{gather}
where $\lambda_{I},\lambda_{D}$ are constant weights and $r_{x} = \frac{1}{\sqrt{\frac{\partial X^{2}}{\partial x} + \frac{\partial Z^{2}}{\partial x} }}$, $r_{y} = \frac{1}{\sqrt{\frac{\partial Y^{2}}{\partial y} + \frac{\partial Z^{2}}{\partial y} }}$. 

As the energy function (Eq.~(\ref{energyfunc})) is based on a linearisation of the 
data term (Eq.~(\ref{dt})) and convex TV regularizer (Eq.~(\ref{eq6})), the energy function 
can be solved using convex solver. An iterative solver can be obtained by 
deriving the energy function (Eq.~(\ref{energyfunc})) as its primal-dual formulation 
and implemented in parallel on GPUs. For more implementation details, the keen 
reader is recommended to read~\cite{jaimezprimal}.

\subsection{Self-Calibration}\label{selfcalibration}

Scene flow computation requires that the RGB and depth data be spatially 
aligned and temporally synchronized. 
The data considered in this paper were captured by Kinect sensors and are 
temporally synchronized. However, the RGB and depth channels may not be 
spatially registered if calibration was not performed properly before recording 
the data. For the RGB-D datasets with spatial misalignment, 
we propose an effective self-calibration method to perform spatial 
alignment without knowledge of the cameras parameters. The alignment is based 
on a pinhole model through which depth maps are transformed into the same view 
of the RGB video. Let $p_i$ be a point in an RGB frame and $p_i'$ be the 
corresponding point in the depth map. The 2D homography mapping 
$H$ satisfying $p_i=Hp_i'$  is a $3\times{}3$ projective transformation for the 
alignment. 
Following the method in~\cite{hartley2003multiple}, we chose a set of 
matching points in an RGB frame and its corresponding depth map. Using four 
pairs of corresponding points, $H$ is obtained 
through direct linear transformation. 
Let $p_i'=(x_i',y_i',1)^T$, $h_j^T$ be the $j_{th}$ row of $H$ and 
$\textbf{0}=[0,0,0]^T$.
The vector cross product equation $p_i\times{}Hp_i'=\textbf{0}$ is written as~\cite{hartley2003multiple}:
\begin{equation}
 \left[\begin{array}{ccc}
        \textbf{0}^T & -p_i^T & y_i'p_i^T\\
        p_i^T & \textbf{0}^T & -x_i'p_i^T
       \end{array}
 \right]\left(\begin{array}{c}
               h_1\\h_2\\h_3 
              \end{array}
 \right)=\textbf{0},
\end{equation}
where the up-to-scale equation is omitted.
A better estimation of $H$ is achieved by minimising (for example, using 
Levenberg-Marquardt algorithm~\cite{kanzow2005withdrawn}) the following 
objective function with more matching points:
\begin{IEEEeqnarray}{c}
\argmin_{\hat{H},\hat{p_i},\hat{p_i}'}\sum_i[d(p_i,\hat{p_i})^2+d(p_i',\hat{p_i}')]\nonumber\\
\text{s.t.}\ \hat{p_i}=\hat{H}\hat{p_i}'\ \text{for}\ \forall{}i\label{ml}
\end{IEEEeqnarray}
 
In Eq.~\eqref{ml}, $d(\cdot)$ is the distance function and $\hat{H}$ is the 
optimal estimation of the homography mapping while $\hat{p_i}$ and $\hat{p_i}'$ are estimated matching points from 
$\{p_i,p_i'\}$. 
Because the process of selecting matching points may not be reliable, the random 
sample consensus~(RANSAC) algorithm is applied to exclude outliers. 
By transforming the depth map using the 2D projective transformation $H$, the RGB video and 
its corresponding depth video are spatially aligned.

\subsection{Construction of Hand-crafted SFAM }\label{sfam}
SFAM encodes a video sample into a single dynamic image to 
take advantage of the available pre-trained models for standard ConvNets 
architecture without training millions of parameters afresh. There are several 
ways to encode the video sequences into dynamic 
images~\cite{bobick2001recognition,
man2006individual,
Yang2012a,pichao2015,pichaoTHMS,bilen2016dynamic}, but how to encode 
the scene flow vectors into one dynamic image still needs to be explored. As 
described in Section~\ref{pdflow}, one scene flow vector $\textbf{s} = 
(\mu,\upsilon,\omega)^{T}$ is obtained by matching/tracking one point in the 
current frame to another in the reference frame; this is one simple Lagrangian 
motion. In order to avoid error in tracking Lagrangian motion over long term, 
we construct SFAM using the Eulerian motion approach and thus, the SFAM 
inherits the merits of both the Eulerian and Lagrangian motion. As we argued 
earlier, the three entries ($\mu,\upsilon,\omega$) in the scene flow vector 
$\textbf{s}$ for each point can be considered as three channels. Hence a
scene flow between two pairs of RGB-D images ($I_{0}$, $Z_{0}$ and $I_{1}$, 
$Z_{1}$) can be reorganized as one three-channel SFM ($X_{\mu}$, 
$X_{\upsilon}$, $X_{\omega}$), and the RGB-D video sequences can be represented 
as SFM sequences. Based on the SFM sequences, there are several ways to 
construct the SFAM. 

\subsubsection{SFAM-D}
Inspired by the construction of Depth Motion Maps (DMM)~\cite{Yang2012a}, 
we accumulate the absolute differences between consecutive SFMs and denote it 
as SFAM-D. It is written as:
\begin{equation}
\text{SFAM-D}_{i}  =  \sum\limits_{t=1}^{T-1} |X_{i}^{t+1} - X_{i}^{t}|  ~~~~      i \in (\mu,\upsilon,\omega),                     
\end{equation}
where $t$ denotes the map number and $T$ represents the total number of maps (the same for the following sections). This representation characterizes the distribution of the accumulated motion difference energy. 

\subsubsection{SFAM-S}
Similarly to SFAM-D, we construct the SFAM-S (S here denotes the sum) by 
accumulating the sum between consecutive SFMs. This can be written as:
\begin{equation}
\text{SFAM-S}_{i}  =  \sum\limits_{t=1}^{T-1} (X_{i}^{t+1} + X_{i}^{t})  ~~~~      i \in (\mu,\upsilon,\omega).                     
\end{equation}
This representation mainly captures the large motion of an action after normalization.

\subsubsection{SFAM-RP}
Inspired by the work reported in~\cite{bilen2016dynamic}, we adopt the rank 
pooling method to encode SFM sequence into one action image. Let 
$X_{1},...,X_{T}$ denote the SFM sequence where each $X_{t}$ contains three 
channels ($X_{\mu}$, $X_{\upsilon}$, $X_{\omega}$), and $\varphi(X_{t}) \in 
\mathbb{R}^{d}$ be a representation or feature vector extracted from each 
individual map, $X_{t}$. Herein, we directly apply rank pooling to the $X$, 
thus, $\varphi(\cdot)$ equals to identity matrix.  Let $V_{t} = 
\dfrac{1}{t}\sum_{\tau=1}^{t}\varphi(X_{\tau})$ be time average of these 
features up to time $t$. The ranking function associates with each time $t$ a 
score $S(t|\textbf{d}) = <\textbf{d}, V_{t}>$, where $\textbf{d} \in 
\mathbb{R}^{d}$ is a vector of parameters. The function parameters $\textbf{d}$ 
are learned so that the scores reflect the order of the maps in the video. In 
general, more recent frames are associated with larger scores, i.e. $~q > t 
\Rightarrow S(q|\textbf{d}) > S(t|\textbf{d})$. Learning $\textbf{d}$ is 
formulated as a convex optimization problem using 
RankSVM~\cite{smola2004tutorial}:

\begin{equation}
\begin{aligned}
\textbf{d}^{*} &=\rho(X_{1},...,X_{T};\varphi) = \mathop{\arg\min}_{\textbf{d}} E(\textbf{d}),\\
  E(\textbf{d}) &= \dfrac{\lambda}{2}\parallel \textbf{d} \parallel^{2} +\\
  & \dfrac{2}{T(T-1)}\times\sum\limits_{q>t}\max\{{0,1-S(q|\textbf{d}) + S(t|\textbf{d})}\}.
\end{aligned}
\end{equation}

The first term in this objective function is the usual quadratic regular term 
used in SVMs. 
The second term is a hinge-loss soft-counting how many pairs $q > t$ are 
incorrectly ranked by the scoring function. Note in particular that a pair is 
considered correctly ranked only if scores are separated by at least a unit 
margin, $i.e.~ S(q|\textbf{d}) > S(t|\textbf{d}) + 1$. 

Optimizing the above equation defines a function $\rho(X_{1},...,X_{T};\varphi)$ 
that maps a sequence of $T$ SFMs to a single vector $\textbf{d}^{*}$. Since 
this vector contains enough information to rank all the frames in the SFM 
sequence, it aggregates information from all of them and can be used as a 
sequence descriptor.  In our work, the rank pooling is applied in a bidirectional 
manner to convert each SFM sequence into two action maps, SFAM-RPf (forward) 
and SFAM-RPb (backward). This representation captures the different 
types of importance associated with frames in one action and assigns more 
weight to recent frames. 

\subsubsection{SFAM-AMRP}

In previous sections, all the three channels are considered as separate 
channels in constructing SFAM. However, the specific relationship (independent 
or otherwise) between them is yet to be ascertained. To study this 
relationship, we adopt a simple method \textit{viz.}, using amplitude of the 
scene flow vector $\textbf{s}$ to represent the relations between the three 
components. Thus, for each triple ($X_{\mu}$, $X_{\upsilon}$, $X_{\omega}$) we 
obtain a new amplitude map, $X_{am}$. Based on the $X_{am} = \sqrt{X_{\mu}^{2} 
+ X_{\upsilon}^{2} + X_{\omega}^{2}}$, the rank pooling method is applied to 
encode the scene flow maps into two action maps, SFAM-AMRPf and SFAM-AMRPb. This 
representation exploits the weights of frames based on the motion magnitude.

\subsubsection{SFAM-LABRP}

To further investigate the relationship amongst the triple ($X_{\mu}$, 
$X_{\upsilon}$,  $X_{\omega}$), they are transformed nonlinearly into another 
space, similarly to the manner of transforming RGB color space to \textit{Lab} 
space. The \textit{Lab} space is designed to approximate the human visual 
system.  Based on these transformed maps, the rank pooling method is applied to 
encode the sequence into two action maps, SFAM-LABRPf and SFAM-LABRPb. 

A few examples of the SFAM variants are shown in Figure~\ref{SFAMs} 
for action ``Bounce Basketball" from M$^{\textbf{2}}$I 
Dataset~\cite{liu2016benchmarking}. It can be seen that different variants of 
SFAM capture and encode SFM sequence into action maps with large visual 
differences.

\begin{figure}[t]
\begin{center}
{\includegraphics[height = 25mm, width = 85mm]{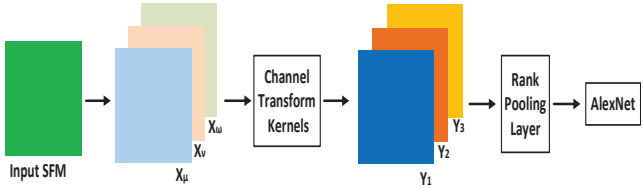}}
\end{center}
\caption{The framework for constructing SFAM with Channel Transform Kernels using ConvNets. }
\label{framework}
\end{figure}

\subsection{Constructing SFAM with Channel Transform Kernels (SFAM-CTKRP)}\label{sfam-learning}

\begin{figure}[t]
\begin{center}
{\includegraphics[height = 25mm, width = 85mm]{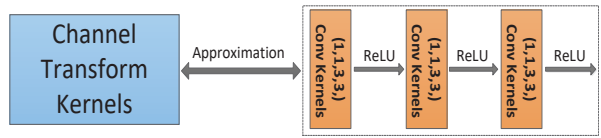}}
\end{center}
\caption{Illustration of approximate computation for Channel Transform Kernels 
using convolution kernels followed by nonlinear transforms. }
\label{approximate}
\end{figure}

In previous sections, we have presented the concept of SFAM and its several 
variants. 
However, it has been empirically observed that none of them can achieve the best 
results for all the datasets or scenarios. One reason adduced for this is 
that during the construction of the SFAM, the relationship amongst the triple 
($X_{\mu}$, $X_{\upsilon}$, $X_{\omega}$) are hand-crafted.  To learn the 
relationship amongst the elements of the triple ($X_{\mu}$, $X_{\upsilon}$, 
$X_{\omega}$) from data with ConvNets, we propose a Channel Transform Kernels 
as follows. 
Let $Y_{1}, Y_{2}, Y_{3}$ be the new learned maps from the original 
triple ($X_{\mu}$, $X_{\upsilon}$, $X_{\omega}$), the relationship between them 
can be formulated as:
\begin{gather}
Y_{1} = \varphi_{1}(\omega_{1}X_{\mu} + \omega_{2}X_{\upsilon} + \omega_{3}X_{\omega}) \notag\\
Y_{2} = \varphi_{2}(\omega_{4}X_{\mu} + \omega_{5}X_{\upsilon} + \omega_{6}X_{\omega}) \\\label{relation} 
Y_{3} = \varphi_{3}(\omega_{7}X_{\mu} + \omega_{8}X_{\upsilon} + \omega_{9}X_{\omega})\notag
\end{gather}
where $Y$ has the same size with $X$, $\omega$ are scalar values and $\varphi$ 
denotes the transforms that need to be learned. The learning framework is 
illustrated in Figure~\ref{framework}. There are different ways to learn these 
Channel Transform Kernels. For sake of simplicity, in this work we approximated 
the transforms by three successive convolution layers, where each layer 
comprises nine convolutional kernels with size $1 \times 1$ and followed by 
ReLU nonlinear transform, as illustrated in Figure~\ref{approximate}. Based on 
RankPool layer~\cite{bilen2016dynamic} for temporal encoding, we can construct 
the SFAM with the proposed Channel Transform Kernels using ConvNets.

\subsection{Multiply-Score Fusion for Classification}
After construction of the several variants of SFAM, we propose to adopt one 
effective late score fusion method, namely, multiply-score fusion method, to 
improve the final recognition accuracy. 
Take SFAM-RP for example,  as illustrated in Figure~\ref{scorefusion}, two 
SFAM-RP, one SFAM-RPf and one SFAM-RPb, are generated for one pair of RGB-D videos 
and they are fed into two different trained ConvNets channels. The score 
vectors output by the two ConvNets are multiplied element-wisely  and 
the max score in the resultant vector is assigned as the probability of the 
test sequence. The index of this max score corresponds to the recognized class 
label. This process can be easily extended into multiple channels.

\section{Experiments}\label{results}

According to the survey of RGB-D datasets~\cite{zhang2016rgb}, we chose two 
public benchmark datasets, which contain both RGB+depth modalities and have 
relatively large training samples to evaluate the proposed method. Specifically 
we chose ChaLearn LAP IsoGD Dataset~\cite{wanchalearn} and 
M$^{\textbf{2}}$I Dataset~\cite{liu2016benchmarking}. In the following, we 
proceed by briefly describing the implementation details and then present 
the experiments and results.

\subsection{Implementation Details}

For scene flow computation, we adopted the public codes provided 
by~\cite{jaimezprimal}.
For rank pooling, we followed the work reported in~\cite{bilen2016dynamic} where 
each channel was generated into one channel dynamic map and then merged 
the three channels into one three-channel map. Differently 
from~\cite{bilen2016dynamic}, we used bidirectional rank pooling. For ChaLearn 
LAP IsoGD Dataset, in order to minimize the interference of the background, it 
is assumed that the background in the histogram of depth maps occupies the last 
peak representing far distances. Specifically, pixels whose depth values are 
greater than a threshold defined by the last peak of the depth histogram minus a 
fixed tolerance (0.1 was set in our experiments) are considered as background 
and removed from the calculation of scene flow by setting their depth values 
to zero. Through this simple process, most of the background can be removed and 
has much contribution to the SFAM.

The AlexNet~\cite{krizhevsky2012imagenet} was adopted in this paper. 
The training procedure of the hand-crafted SFAMs was similar to that described 
in~\cite{krizhevsky2012imagenet}. The network weights were learned using 
the mini-batch stochastic gradient descent with the momentum set to 0.9 
and weight decay set to 0.0005. All hidden weight layers used the 
rectification (RELU) activation function. At each iteration, a mini-batch of 256 
samples was constructed by sampling 256 shuffled training samples. All the images 
were resized to 256 $\times$ 256. The learning rate was set to $10^{-3}$ for 
fine-tuning with pre-trained models on ILSVRC-2012, and then it was decreased 
according to a fixed schedule, which was kept the same for all training sets. 
Different datasets underwent different iterations according to their number of 
training samples. For all experiments, the dropout regularization ratio was set 
to 0.5 in order to reduce complex co-adaptations of neurons in the nets. The 
implementation was derived from the publicly available Caffe 
toolbox~\cite{jia2014caffe} based on one {NVIDIA Tesla K40 GPU} card. Unless 
otherwise specified, all the networks were initialized with the models trained 
over ImageNet~\cite{krizhevsky2012imagenet}. For SFAM-CTKRP, we revised the 
codes of paper~\cite{bilen2016dynamic} based on 
MatConvNet~\cite{vedaldi15matconvnet}. The multiply score fusion method is compared with the other two commonly used late score fusion methods, average  and maximum score fusion on both datasets. This verifies that the SFAMs are likely to be statistically independent and provide complementary information. 

\subsection{ChaLearn LAP IsoGD Dataset} The ChaLearn LAP IsoGD 
Dataset~\cite{wanchalearn} includes 
47933 RGB-D depth sequences, each RGB-D video representing one gesture instance. 
There are 249 gestures performed by 21 different individuals. This dataset does 
not provide the true depth values in their depth videos. To use this dataset for 
scene flow calculation, we estimate the depth values using the average minimum 
and maximum values provided for CGD dataset.  The dataset is divided into 
training, validation and test sets. As the test set is not available 
for public usage, we report the results on the validation set. For this dataset 
the training underwent 25K iterations and the learning rate decreased every 10K 
iterations.

\textbf{Results.} Table~\ref{table1} shows the results of six variants of 
SFAM, 
and compares them with methods in the 
literature~\cite{pami16Jun,wanchalearn,bilen2016dynamic,pichaoTHMS}. Among these 
methods, MFSK combined 3D SMoSIFT~\cite{wan20143d} with (HOG, HOF and 
MBH)~\cite{wang2013action} descriptors.  MFSK+DeepID further included Deep 
hidden IDentity (Deep ID) feature~\cite{sun2014deep}. Thus, these two methods 
utilized not only hand-crafted features but also deep learning features. 
Moreover, they extracted features from RGB and depth separately, 
concatenated them together, and adopted Bag-of-Words (BoW) model as the final 
video representation. The other methods, 
WHDMM+SDI~\cite{pichaoTHMS,bilen2016dynamic}, extracted features and conducted 
classification with ConvNets from depth and RGB individually and adopted 
multiply-score fusion for final recognition.

Compared with these methods, the proposed SFAM outperformed all of them 
significantly. It is worth noting that all the depth values used in the proposed SFAM were estimated 
rather than the exact real depth values. Despite the possible estimation 
errors, our method still achieved promising results.
Interestingly, the proposed variants of SFAM are complementary to each other and 
can improve each other largely by using multiply-score fusion.  Even though 
this dataset is large, on average  144 video clips per class, it is still much 
smaller compared with 1200 images per class in ImageNet. Thus, directly 
training from scratch cannot compete with fine-tuning the trained 
models over ImageNet and this is evident in the results reported in 
Table~\ref{table1}.
By comparing different types of SFAM, we can see that the simple SFAM-S method 
achieved the best results among all types of hand-designed SFAMs. Due to the 
relatively large training data, SFAM-CTKRP achieved the best result among all 
the variants, even though the approximate rank pooling in the 
work reported in~\cite{bilen2016dynamic} was shown to be worse than rank 
pooling solved by RankSVM~\cite{smola2004tutorial}. The reasons for these two 
phenomenona probably are as follows: under the inaccurate estimation of the 
depth values, the scene flow computation will be affected and based on this 
inaccurate scene flow vectors, rank pooling can not achieve its full efficacy. 
In other words, the rank pooling method is sensitive to noise. Instead, the 
proposed Channel Transform Kernels cannot only exploit the relations amongst 
the channels but also decrease the effects of noise after channel transforms. 

\begin{table}[!ht]
\centering
\begin{tabular}{|c|c|}
\hline
Method & Accuracy \\
\hline
MFSK~\cite{pami16Jun,wanchalearn}  & 18.65\%\\
\hline
MFSK+DeepID~\cite{pami16Jun,wanchalearn}  & 18.23\%\\
\hline
SDI~\cite{bilen2016dynamic}  & 20.83\%\\
\hline
WHDMM~\cite{pichaoTHMS}  & 25.10\%\\
\hline
WHDMM+SDI~\cite{pichaoTHMS,bilen2016dynamic}  & 25.52\%\\
\hline\hline
SFAM-D (training from scratch) & 9.23\%\\
\hline
SFAM-D & 18.86\%\\
\hline
SFAM-S (training from scratch)  & 18.10\%\\
\hline
SFAM-S  & 25.83\%\\
\hline
SFAM-RP  & 23.62\%\\
\hline
SFAM-AMRP  & 18.21\%\\
\hline
SFAM-LABRP  & 23.35\%\\
\hline
SFAM-CTKRP  & 27.48\%\\
\hline
Max-Score Fusion All  & 33.24\%\\
\hline
Average-Score Fusion All  & 34.86\%\\
\hline
Multiply-Score Fusion All  & \textbf{36.27\%}\\
\hline
\end{tabular}
\caption{Results and Comparison on the ChaLearn LAP IsoGD Dataset. \label{table1}}
\end{table}

%

\subsection{M$^{\textbf{2}}$I Dataset}
Multi-modal \& Multi-view \& Interactive (M$^{\textbf{2}}$I) 
Dataset~\cite{liu2016benchmarking} 
provides person-person interaction 
actions and person-object interaction actions. It contains both the 
front and side views; denoted as Front View (FV) and Side View (SV). It 
consists of 22 action categories and a total of 22 unique individuals. Each 
action was performed twice by 20 groups (two persons in a group). In total, 
M$^{\textbf{2}}$I dataset contains 1760 samples (22 actions $\times$ 20 groups 
$\times$ 2 views $\times$ 2 run).  For evaluation, all samples were divided with respect to the groups into 
a training set (8 groups), a validation set (6 groups) and a test set (6 
groups). The final action recognition results are obtained with the test set. 
For this dataset the training underwent 6K iterations and the learning rate 
decreased every 3K iterations.
 
\textbf{Results.} We followed the experimental settings as 
in~\cite{liu2016benchmarking} 
and compared the results on two scenarios: single task scenario and cross-view 
scenario. 
 The baseline methods were based on iDT features~\cite{wang2013action}  
generated from optical flow and has been shown to be very effective in 2D 
action recognition. Specifically, for the BoW framework, a set of local 
spatio-temporal features were extracted, including iDT-Tra, iDT-HOG, iDT-HOF, 
iDT-MBH, iDT-HOG+HOF, iDT-HOF+MBH and iDT-COM (concatenation of all 
descriptors); for fisher vector framework, they only used the iDT-COM feature 
for evaluation. For comparisons, we only show several best results achieved by 
baseline methods for each scenario. Table~\ref{table2} shows the comparisons on 
the M$^{\textbf{2}}$I Dataset for single task scenario, that is, learning and 
testing in the same view while Table~\ref{table3} presents the comparisons for 
cross-view scenario. Due to the lack of training data, SFAM-CTKRP could not 
converge steadily and the results varied largely, thus, we did not show its 
results. For this dataset, SFAM-AMRP achieved the best result for side view 
while SFAM-LABRP achieved the best result for front view. From 
Table~\ref{table2} we can see that for scene flow estimation based on real true 
depth values, the rank pooling-based method achieved better results than SFAM-D 
and SFAM-S, which are consistent with the conclusion 
in~\cite{liu2016benchmarking}. SFAM-AMRP achieved the best results for two 
cross-view scenarios which can be seen from Table~\ref{table3}.  Interestingly, 
even though our proposed SFAM  did not solve any transfer learning problem as 
in~\cite{liu2016benchmarking} but directly training with the side/front view 
and testing in the front/side view, it still outperformed the best baseline 
method significantly, especially in the SV $\rightarrow$ FV setting. This bonus 
advantage reflects the effectiveness of proposed method. 

\begin{table}[!ht]
\centering
\begin{tabular}{|c|c|c|} \hline 
\multirow{2}{*}{Method} 
& \multicolumn{2}{c|}{Accuracy}\\
 \cline{2-3}   
 & SV   & FV   \\ \hline
   iDT-Tra (BoW)~\cite{liu2016benchmarking} &  69.8\%  & 65.8\% \\ \hline
   iDT-COM (BoW)~\cite{liu2016benchmarking} &  76.9\%  & 75.3\% \\ \hline
   iDT-COM (FV)~\cite{liu2016benchmarking} &  80.7\%  & 79.5\% \\ \hline
   iDT-MBH (BoW)~\cite{liu2016benchmarking} & 77.2\%  & 79.6\% \\ \hline\hline
   SFAM-D & 71.2\%  & 83.0\% \\ \hline
   SFAM-S & 70.1\%  & 75.0\% \\ \hline
   SFAM-RP & 79.9\%  & 81.8\% \\ \hline
   SFAM-AMRP & 82.2\%  & 78.0\% \\ \hline
   SFAM-LABRP & 72.0\%  & 83.7\% \\ \hline
   Max-Score Fusion All & 87.6\%  & 88.8\% \\ \hline
   Average-Score Fusion All & 88.2\%  & 89.1\% \\ \hline
   Multiply-Score Fusion All & \textbf{89.4\%}  & \textbf{91.2\%} \\ \hline
\end{tabular}
\caption{Comparison on the M$^{\textbf{2}}$I Dataset for single task scenario 
(learning and testing in the same view). \label{table2}}
\end{table}

\begin{table}[!ht]
\centering
\begin{tabular}{|c|c|c|} \hline 
\multirow{2}{*}{Method} 
& \multicolumn{2}{c|}{Accuracy}\\
 \cline{2-3}   
 & SV $\rightarrow$ FV  & FV $\rightarrow$ SV  \\ \hline
   iDT-Tra~\cite{liu2016benchmarking} & 43.3\%  & 39.2\% \\ \hline
   iDT-COM~\cite{liu2016benchmarking} & 70.2\%  & 67.7\% \\ \hline
   iDT-HOG+MBH~\cite{liu2016benchmarking} & 75.8\%  & 71.8\% \\ \hline
   iDT-HOG+HOF~\cite{liu2016benchmarking} & 78.2\%  & 72.1\% \\ \hline\hline
   SFAM-D & 66.7\%  & 65.2\% \\ \hline
   SFAM-S & 68.2\%  & 60.2\% \\ \hline
   SFAM-RP & 71.6\%  & 65.2\% \\ \hline
   SFAM-AMRP & 77.7\%  & 66.7\% \\ \hline
   SFAM-LABRP & 76.9\%  & 65.9\% \\ \hline
   Max-Score Fusion All & 84.7\%  & 73.8\% \\ \hline
   Average-Score Fusion All & 85.3\%  & 75.3\% \\ \hline
   Multiply-Score Fusion All & \textbf{87.6\%}  & \textbf{76.5\%} \\ \hline
\end{tabular}
\caption{Comparison on the M$^{\textbf{2}}$I Dataset 
for cross-view scenario.(SV $\rightarrow$ FV: learning in the side view and 
test in the front view; FV $\rightarrow$ SV: learning in the front view and 
testing in the side view.) \label{table3}}
\end{table}

\section{Conclusion and Future Work}\label{conclusion}
We propose a novel method for action recognition based on scene flow. 
In particular, scene flow vectors are estimated from registered RGB and depth 
data. A new representation based on scene flow vectors, SFAM, and several 
variants that capture the spatio-temporal information from different 
perspectives are proposed for 3D action recognition. In order to exploit the 
relationships amongst the three channels of scene flow map, we propose to learn 
the Channel Transform Kernels, end-to-end, with  ConvNets from data. 
Experiments on two benchmark datasets have demonstrated the effectiveness of the 
proposed method. For the future work, we will improve the temporal encoding 
method based on scene flow vectors.

\section*{Acknowledgement}
The authors would like to thank NVIDIA Corporation for the donation of a Tesla K40 GPU card used in this research.



\small 
\begin{thebibliography}{10}\itemsep=-1pt

\bibitem{bilen2016dynamic}
H.~Bilen, B.~Fernando, E.~Gavves, A.~Vedaldi, and S.~Gould.
\newblock Dynamic image networks for action recognition.
\newblock In {\em CVPR}, 2016.

\bibitem{bobick2001recognition}
A.~F. Bobick and J.~W. Davis.
\newblock The recognition of human movement using temporal templates.
\newblock {\em IEEE Transactions on pattern analysis and machine intelligence},
  23(3):257--267, 2001.

\bibitem{donahue2015long}
J.~Donahue, L.~Anne~Hendricks, S.~Guadarrama, M.~Rohrbach, S.~Venugopalan,
  K.~Saenko, and T.~Darrell.
\newblock Long-term recurrent convolutional networks for visual recognition and
  description.
\newblock In {\em CVPR}, pages 2625--2634, 2015.

\bibitem{du2015hierarchical}
Y.~Du, W.~Wang, and L.~Wang.
\newblock Hierarchical recurrent neural network for skeleton based action
  recognition.
\newblock In {\em CVPR}, pages 1110--1118, 2015.

\bibitem{fernando2016rank}
B.~Fernando, S.~Gavves, O.~Mogrovejo, J.~Antonio, A.~Ghodrati, and
  T.~Tuytelaars.
\newblock Rank pooling for action recognition.
\newblock {\em IEEE Transactions on Pattern Analysis and Machine Intelligence},
  2016.

\bibitem{hadfield2014scene}
S.~Hadfield and R.~Bowden.
\newblock Scene particles: Unregularized particle-based scene flow estimation.
\newblock {\em IEEE transactions on pattern analysis and machine intelligence},
  36(3):564--576, 2014.

\bibitem{hartley2003multiple}
R.~Hartley and A.~Zisserman.
\newblock {\em Multiple view geometry in computer vision}.
\newblock Cambridge university press, 2003.

\bibitem{hornacek2014sphereflow}
M.~Hornacek, A.~Fitzgibbon, and C.~Rother.
\newblock Sphereflow: 6 {DoF} scene flow from {RGB-D} pairs.
\newblock In {\em CVPR}, pages 3526--3533, 2014.

\bibitem{hou2016skeleton}
Y.~Hou, Z.~Li, P.~Wang, and W.~Li.
\newblock Skeleton optical spectra based action recognition using convolutional
  neural networks.
\newblock {\em IEEE Transactions on Circuits and Systems for Video Technology},
  2016.

\bibitem{hu2015jointly}
J.-F. Hu, W.-S. Zheng, J.~Lai, and J.~Zhang.
\newblock Jointly learning heterogeneous features for {RGB-D} activity
  recognition.
\newblock In {\em CVPR}, pages 5344--5352, 2015.

\bibitem{jaimezprimal}
M.~Jaimez, M.~Souiai, J.~Gonzalez-Jimenez, and D.~Cremers.
\newblock A primal-dual framework for real-time dense {RGB-D} scene flow.
\newblock In {\em ICRA}, pages 98--104, 2015.

\bibitem{ji20133d}
S.~Ji, W.~Xu, M.~Yang, and K.~Yu.
\newblock {3D} convolutional neural networks for human action recognition.
\newblock {\em Pattern Analysis and Machine Intelligence, IEEE Transactions
  on}, 35(1):221--231, 2013.

\bibitem{jia2016low}
C.~Jia and Y.~Fu.
\newblock Low-rank tensor subspace learning for rgb-d action recognition.
\newblock {\em IEEE Transactions on Image Processing}, 25(10):4641--4652, 2016.

\bibitem{jia2014caffe}
Y.~Jia, E.~Shelhamer, J.~Donahue, S.~Karayev, J.~Long, R.~B. Girshick,
  S.~Guadarrama, and T.~Darrell.
\newblock Caffe: Convolutional architecture for fast feature embedding.
\newblock In {\em Proc. ACM international conference on Multimedia (ACM MM)},
  pages 675--678, 2014.

\bibitem{kanzow2005withdrawn}
C.~Kanzow, N.~Yamashita, and M.~Fukushima.
\newblock Withdrawn: Levenberg--marquardt methods with strong local convergence
  properties for solving nonlinear equations with convex constraints.
\newblock {\em Journal of Computational and Applied Mathematics},
  173(2):321--343, 2005.

\bibitem{Kong2015CVPR}
Y.~Kong and Y.~Fu.
\newblock Bilinear heterogeneous information machine for {RGB-D} action
  recognition.
\newblock In {\em CVPR}, pages 1054--1062, 2015.

\bibitem{krizhevsky2012imagenet}
A.~Krizhevsky, I.~Sutskever, and G.~E. Hinton.
\newblock Imagenet classification with deep convolutional neural networks.
\newblock In {\em Proc. Annual Conference on Neural Information Processing
  Systems (NIPS)}, pages 1106--1114, 2012.

\bibitem{lan2015beyond}
Z.~Lan, M.~Lin, X.~Li, A.~G. Hauptmann, and B.~Raj.
\newblock Beyond gaussian pyramid: Multi-skip feature stacking for action
  recognition.
\newblock In {\em CVPR}, pages 204--212, 2015.

\bibitem{pichaoSPL}
C.~Li, Y.~Hou, P.~Wang, and W.~Li.
\newblock Joint distance maps based action recognition with convolutional
  neural network.
\newblock {\em IEEE Signal Processing Letters}, 2017.

\bibitem{li2010action}
W.~Li, Z.~Zhang, and Z.~Liu.
\newblock Action recognition based on a bag of {3D} points.
\newblock In {\em CVPRW}, pages 9--14, 2010.

\bibitem{liu2016benchmarking}
A.-A. Liu, N.~Xu, W.-Z. Nie, Y.-T. Su, Y.~Wong, and M.~Kankanhalli.
\newblock Benchmarking a multimodal and multiview and interactive dataset for
  human action recognition.
\newblock {\em IEEE Transactions on cybernetics}, 2016.

\bibitem{liu2016spatio}
J.~Liu, A.~Shahroudy, D.~Xu, and G.~Wang.
\newblock Spatio-temporal {LSTM} with trust gates for {3D} human action
  recognition.
\newblock In {\em Proc. European Conference on Computer Vision}, pages
  816--833, 2016.

\bibitem{junliu2017}
J.~Liu and G.~Wang.
\newblock Global context-aware attention lstm networks for 3d action
  recognition.
\newblock In {\em CVPR}, 2017.

\bibitem{lurange}
C.~Lu, J.~Jia, and C.-K. Tang.
\newblock Range-sample depth feature for action recognition.
\newblock In {\em CVPR}, pages 772--779, 2014.

\bibitem{man2006individual}
J.~Man and B.~Bhanu.
\newblock Individual recognition using gait energy image.
\newblock {\em IEEE transactions on pattern analysis and machine intelligence},
  28(2):316--322, 2006.

\bibitem{menze2015object}
M.~Menze and A.~Geiger.
\newblock Object scene flow for autonomous vehicles.
\newblock In {\em CVPR}, pages 3061--3070, 2015.

\bibitem{ni2015pose}
B.~Ni, P.~Moulin, and S.~Yan.
\newblock Pose adaptive motion feature pooling for human action analysis.
\newblock {\em International Journal of Computer Vision}, 111(2):229--248,
  2015.

\bibitem{Nie2015}
S.~Nie, Z.~Wang, and Q.~Ji.
\newblock A generative restricted boltzmann machine based method for
  high-dimensional motion data modeling.
\newblock {\em Computer Vision and Image Understanding}, pages 14--22, 2015.

\bibitem{Oreifej2013}
O.~Oreifej and Z.~Liu.
\newblock {HON4D}: Histogram of oriented {4D} normals for activity recognition
  from depth sequences.
\newblock In {\em CVPR}, pages 716--723, 2013.

\bibitem{peng2016bag}
X.~Peng, L.~Wang, X.~Wang, and Y.~Qiao.
\newblock Bag of visual words and fusion methods for action recognition:
  Comprehensive study and good practice.
\newblock {\em Computer Vision and Image Understanding}, 2016.

\bibitem{peng2014action}
X.~Peng, C.~Zou, Y.~Qiao, and Q.~Peng.
\newblock Action recognition with stacked fisher vectors.
\newblock In {\em ECCV}, pages 581--595, 2014.

\bibitem{quiroga2014dense}
J.~Quiroga, T.~Brox, F.~Devernay, and J.~Crowley.
\newblock Dense semi-rigid scene flow estimation from {RGBD} images.
\newblock In {\em ECCV}, pages 567--582. 2014.

\bibitem{rahmani2015learning}
H.~Rahmani and A.~Mian.
\newblock Learning a non-linear knowledge transfer model for cross-view action
  recognition.
\newblock In {\em CVPR}, pages 2458--2466, 2015.

\bibitem{shahroudy2016ntu}
A.~Shahroudy, J.~Liu, T.-T. Ng, and G.~Wang.
\newblock {NTU RGB+ D}: A large scale dataset for {3D} human activity analysis.
\newblock In {\em CVPR}, 2016.

\bibitem{simonyan2014two}
K.~Simonyan and A.~Zisserman.
\newblock Two-stream convolutional networks for action recognition in videos.
\newblock In {\em NIPS}, pages 568--576, 2014.

\bibitem{smola2004tutorial}
A.~J. Smola and B.~Sch{\"o}lkopf.
\newblock A tutorial on support vector regression.
\newblock {\em Statistics and computing}, 14(3):199--222, 2004.

\bibitem{sun2015layered}
D.~Sun, E.~B. Sudderth, and H.~Pfister.
\newblock Layered {RGBD} scene flow estimation.
\newblock In {\em CVPR}, pages 548--556, 2015.

\bibitem{sun2014deep}
Y.~Sun, X.~Wang, and X.~Tang.
\newblock Deep learning face representation from predicting 10,000 classes.
\newblock In {\em CVPR}, 2014.

\bibitem{tran2015learning}
D.~Tran, L.~Bourdev, R.~Fergus, L.~Torresani, and M.~Paluri.
\newblock Learning spatiotemporal features with {3D} convolutional networks.
\newblock In {\em ICCV}, pages 4489--4497, 2015.

\bibitem{vedaldi15matconvnet}
A.~Vedaldi and K.~Lenc.
\newblock Matconvnet -- convolutional neural networks for matlab.
\newblock In {\em Proceeding of the {ACM} Int. Conf. on Multimedia}, 2015.

\bibitem{vedula2005three}
S.~Vedula, P.~Rander, R.~Collins, and T.~Kanade.
\newblock Three-dimensional scene flow.
\newblock {\em Pattern Analysis and Machine Intelligence, IEEE Transactions
  on}, 27(3):475--480, 2005.

\bibitem{veeriah2015differential}
V.~Veeriah, N.~Zhuang, and G.-J. Qi.
\newblock Differential recurrent neural networks for action recognition.
\newblock In {\em ICCV}, pages 4041--4049, 2015.

\bibitem{vemulapalli2014human}
R.~Vemulapalli, F.~Arrate, and R.~Chellappa.
\newblock Human action recognition by representing {3D} skeletons as points in
  a lie group.
\newblock In {\em CVPR}, pages 588--595, 2014.

\bibitem{vogel2013piecewise}
C.~Vogel, K.~Schindler, and S.~Roth.
\newblock Piecewise rigid scene flow.
\newblock In {\em ICCV}, pages 1377--1384, 2013.

\bibitem{pami16Jun}
J.~Wan, G.~Guo, and S.~Z. Li.
\newblock Explore efficient local features from {RGB-D} data for one-shot
  learning gesture recognition.
\newblock {\em IEEE Transactions on Pattern Analysis and Machine Intelligence},
  38(8):1626--1639, Aug 2016.

\bibitem{wanchalearn}
J.~Wan, S.~Z. Li, Y.~Zhao, S.~Zhou, I.~Guyon, and S.~Escalera.
\newblock Chalearn looking at people {RGB-D} isolated and continuous datasets
  for gesture recognition.
\newblock In {\em Proc. IEEE Computer Society Conference on Computer Vision and
  Pattern Recognition Workshops (CVPRW)}, pages 1--9, 2016.

\bibitem{wan20143d}
J.~Wan, Q.~Ruan, W.~Li, G.~An, and R.~Zhao.
\newblock 3d smosift: three-dimensional sparse motion scale invariant feature
  transform for activity recognition from rgb-d videos.
\newblock {\em Journal of Electronic Imaging}, 23(2), 2014.

\bibitem{wang2013action}
H.~Wang and C.~Schmid.
\newblock Action recognition with improved trajectories.
\newblock In {\em ICCV}, pages 3551--3558, 2013.

\bibitem{wang2012mining}
J.~Wang, Z.~Liu, Y.~Wu, and J.~Yuan.
\newblock Mining actionlet ensemble for action recognition with depth cameras.
\newblock In {\em CVPR}, pages 1290--1297, 2012.

\bibitem{wang2015action}
L.~Wang, Y.~Qiao, and X.~Tang.
\newblock Action recognition with trajectory-pooled deep-convolutional
  descriptors.
\newblock In {\em CVPR}, pages 4305--4314, 2015.

\bibitem{pichao2015}
P.~Wang, W.~Li, Z.~Gao, C.~Tang, J.~Zhang, and P.~O. Ogunbona.
\newblock Convnets-based action recognition from depth maps through virtual
  cameras and pseudocoloring.
\newblock In {\em ACM MM}, pages 1119--1122, 2015.

\bibitem{pichaoTHMS}
P.~Wang, W.~Li, Z.~Gao, J.~Zhang, C.~Tang, and P.~Ogunbona.
\newblock Action recognition from depth maps using deep convolutional neural
  networks.
\newblock {\em Human-Machine Systems, IEEE Transactions on}, 46(4):498--509,
  2016.

\bibitem{pichaoicprwb}
P.~Wang, W.~Li, S.~Liu, Z.~Gao, C.~Tang, and P.~Ogunbona.
\newblock Large-scale isolated gesture recognition using convolutional neural
  networks.
\newblock In {\em Proceedings of ICPRW}, 2016.

\bibitem{pichaoicprwa}
P.~Wang, W.~Li, S.~Liu, Y.~Zhang, Z.~Gao, and P.~Ogunbona.
\newblock Large-scale continuous gesture recognition using convolutional neural
  networks.
\newblock In {\em Proceedings of ICPRW}, 2016.

\bibitem{pichao2014}
P.~Wang, W.~Li, P.~Ogunbona, Z.~Gao, and H.~Zhang.
\newblock Mining mid-level features for action recognition based on effective
  skeleton representation.
\newblock In {\em DICTA}, 2014.

\bibitem{pichao2016}
P.~Wang, Z.~Li, Y.~Hou, and W.~Li.
\newblock Action recognition based on joint trajectory maps using convolutional
  neural networks.
\newblock In {\em ACM MM}, pages 102--106, 2016.

\bibitem{wedel2011stereoscopic}
A.~Wedel, T.~Brox, T.~Vaudrey, C.~Rabe, U.~Franke, and D.~Cremers.
\newblock Stereoscopic scene flow computation for {3D} motion understanding.
\newblock {\em International Journal of Computer Vision}, 95(1):29--51, 2011.

\bibitem{wu2015watch}
C.~Wu, J.~Zhang, S.~Savarese, and A.~Saxena.
\newblock Watch-n-patch: Unsupervised understanding of actions and relations.
\newblock In {\em CVPR}, pages 4362--4370, 2015.

\bibitem{yangsuper}
X.~Yang and Y.~Tian.
\newblock Super normal vector for activity recognition using depth sequences.
\newblock In {\em CVPR}, pages 804--811, 2014.

\bibitem{Yang2012a}
X.~Yang, C.~Zhang, and Y.~Tian.
\newblock Recognizing actions using depth motion maps-based histograms of
  oriented gradients.
\newblock In {\em ACM MM}, pages 1057--1060, 2012.

\bibitem{yu2016structure}
M.~Yu, L.~Liu, and L.~Shao.
\newblock Structure-preserving binary representations for rgb-d action
  recognition.
\newblock {\em IEEE transactions on pattern analysis and machine intelligence},
  38(8):1651--1664, 2016.

\bibitem{yue2015beyond}
J.~Yue-Hei~Ng, M.~Hausknecht, S.~Vijayanarasimhan, O.~Vinyals, R.~Monga, and
  G.~Toderici.
\newblock Beyond short snippets: Deep networks for video classification.
\newblock In {\em CVPR}, pages 4694--4702, 2015.

\bibitem{zhang2016rgb}
J.~Zhang, W.~Li, P.~O. Ogunbona, P.~Wang, and C.~Tang.
\newblock {RGB-D}-based action recognition datasets: A survey.
\newblock {\em Pattern Recognition}, 60:86--105, 2016.

\bibitem{zhang20013d}
Y.~Zhang and C.~Kambhamettu.
\newblock On 3d scene flow and structure estimation.
\newblock In {\em CVPR}, pages 778--785, 2001.

\end{thebibliography}
\end{document}